\PassOptionsToPackage{numbers, compress}{natbib}

\documentclass{article}
\usepackage[final]{neurips_2025}

\usepackage[utf8]{inputenc} 
\usepackage[T1]{fontenc}    
\usepackage{hyperref}       
\usepackage{url}            
\usepackage{booktabs}       
\usepackage{amsfonts}       
\usepackage{nicefrac}       
\usepackage{microtype}      
\usepackage{xcolor}         
\usepackage{graphicx}
\usepackage{float}
\usepackage{tabularx}
\usepackage{subcaption}
\definecolor{promptbg}{RGB}{248,248,248}
\definecolor{promptborder}{RGB}{200,200,200}

\usepackage[most]{tcolorbox}
\tcbuselibrary{listings,breakable}
\tcbset{
  colback=promptbg,
  colframe=promptborder,
  boxrule=0.6pt,
  arc=2mm,
  left=1em,right=1em,top=0.8em,bottom=0.8em,
  breakable
}
\newtcblisting{promptbox}[1][]{listing only, title=#1, listing options={
    basicstyle=\ttfamily\scriptsize,
    breaklines=true,
    columns=fullflexible,
    keepspaces=true,
    showstringspaces=false,
  }
}
\newcolumntype{Y}{>{\raggedright\arraybackslash}X}

\makeatletter
\providecommand{\workshoptitle}[1]{Foundations of Reasoning in Language Models}
\makeatother
\title{Peek-a-Boo Reasoning: Contrastive Region Masking in MLLMs}

\author{
  Isha Chaturvedi \\
  Independent Researcher \\
  \texttt{chaturvedi.isha6@gmail.com}
  \And
  Anjana Nair \\
  Algoverse AI Research \\
  \texttt{nairanjana696@gmail.com}
  \And
  Yushen Li \\
  Algoverse AI Research \\
  \texttt{ethanli9688@outlook.com}
  \And
  Adhitya Rajendra Kumar \\
  Algoverse AI Research \\
  \texttt{email2adhitya@gmail.com}
  \And
  Kevin Zhu \\
  Algoverse AI Research \\
  \texttt{zhu502846@berkeley.edu}
  \And
  Sunishchal Dev \\
  Algoverse AI Research \\
  \texttt{sunishchaldev@gmail.com}
  \And
  Ashwinee Panda \\
  Princeton University \\
  \texttt{ashwinee@princeton.edu}
  \And
  Vasu Sharma \\
  Algoverse AI Research \\
  \texttt{sharma.vasu55@gmail.com}
} 
\begin{document}

\maketitle
{
\renewcommand{\thefootnote}{}\footnotetext{
\hspace{-1.5mm}*Accepted at the NeurIPS 2025 Workshop on Foundations of Reasoning in Language Models (FoRLM)
}
\renewcommand{\thefootnote}{\arabic{footnote}}
}

\begin{abstract}
We introduce \textbf{Contrastive Region Masking (CRM)}, a training free diagnostic that reveals how multimodal large language models (MLLMs) depend on specific visual regions at each step of chain-of-thought (CoT) reasoning. Unlike prior approaches limited to final answers or attention maps, CRM provides causal, step-level attribution by systematically masking annotated regions and contrasting the resulting reasoning traces with unmasked baselines. Applied to datasets such as \textbf{VisArgs}\cite{zhang2024selective}, CRM reveals distinct failure modes: some models preserve reasoning structure, but hallucinate when evidence is missing, while others ground tightly to visual cues yet collapse under perturbations. By shifting the evaluation from correctness of answers to faithfulness of reasoning, CRM reframes visual benchmarks as diagnostic tools, highlighting the need for multimodal evaluation frameworks that measure not just performance, but also robustness and fidelity of reasoning. 
\end{abstract}

\section{Introduction}

Chain-of-thought reasoning \cite{wei2022chainofthought} has emerged as a powerful technique for improving reasoning in large language models (LLMs). It encourages models to generate step-by-step intermediate reasoning before producing final answers. 
Multimodal large language models (MLLMs) \cite{caffagni2024revolution}\cite{dang2024explainable}\cite{han2025unified}\cite{jin2024efficient}\cite{yin2023survey} have advanced the integration of visual and textual information, enabling complex reasoning over images and language\cite{zhou2023multimodalcot}. However, these models often lack fine-grained selectivity, leading to reasoning failures such as hallucinated logic, semantic drift, and brittle answers when irrelevant or distracting visual content is present\cite{kang2025visualsink}. MLLMs often fail to reason selectively over the most relevant image regions, especialy in complex tasks that involve multiple objects or scenes\cite{wu2023visualcot}. Our analysis on VisArgs dataset shows that state-of-the-art models like GPT-4o \cite{openai2024gpt4oz}, Gemini-1.5-Flash \cite{team2024gemini15}, Qwen-2.5-VL-7b-Instruct \cite{bai2023qwen}, Llama-3.2-90B-Vision-Instruct \cite{meta2024llama3} exhibit these weaknesses.

Prior methods have been fragmented: some inspect attention (e.g., FOCUS \cite{zhong2025focus}), 
some explore grounding or region replay (e.g., VGR \cite{wang2025vgr}, Argus \cite{man2025argus}), 
and others fine-tune visual reasoning (e.g., ICoV \cite{huang2025implicitvision}). 
Crucially, none systematically link step-wise reasoning failures to specific visual regions, 
leaving an important gap in understanding \textbf{how} visual content drives reasoning.


This raises a central question: \textbf{What happens to a model’s reasoning process when key visual regions are removed?}

To address this, we introduce a novel and training-free framework called CRM (Chain of Thought Region Manipulation), as shown in Figure \ref{architecture}, that evaluates MLLMs’ reliance on specific visual regions during reasoning, not only for final answers but for each step in the chain-of-thought (CoT). Our contributions are threefold:

\begin{enumerate}
    \item \textbf{Behavioral Attribution of Visual Reasoning via Contrastive Region Masking (CRM):} We systematically remove ground-truth bounding boxes and track how each CoT step changes, enabling causal attribution between visual regions and reasoning steps. This approach goes beyond saliency maps or final-answer metrics to reveal step-level reasoning failures.
    \item \textbf{Step-to-Region Causality Without Training:} Our framework is black-box and model-agnostic, requiring no retraining. By combining simple region perturbations with semantic similarity over CoT traces, we expose where MLLMs rely on visual evidence and where reasoning fails or hallucinates when cues are absent.
    \item \textbf{Repurposing Structured Benchmarks for Step-Level Diagnostics:} We reframe VisArgs, a dataset with region-level and reasoning annotations as a diagnostic tool. By selectively masking regions associated with each CoT step and measuring resulting reasoning disruptions, we provide insights into model robustness, interpretability, and visual faithfulness.

\end{enumerate}
 

\section{Related Works}
Prior research on evaluating and improving visual reasoning in multimodal large language models (MLLMs) is centered around four thematic axes. One key axis is Grounding Approaches, where methods such as  VGR\cite{wang2025vgr}, Visual Chain-of-Thought(Visual CoT), and Argus focus on grounding visual evidence during reasoning. VGR predicts bounding-box tokens during CoT inference to localize relevant regions, but its reliance on explicit annotations and handcrafted data limits scalability. Visual CoT extends textual reasoning to multimodal settings by prompting models to verbalize intermediate steps that mix linguistic and perceptual information. These approaches emphasize the plausibility of reasoning traces, yet often overlook whether steps are causally supported by the underlying visual input. Argus integrates grounded CoT with visual region grounding to enhance multimodal inference, though its dependence on the limited-scale VisCoT dataset restricts generalization. In contrast, CRM probes causal dependencies directly by systematically masking visual evidence and observing how reasoning steps change.

    The second theme is Training-Free Probes, with methods like FOCUS, which selects regions in a training-free manner using MLLMs to inspect attention. While it provides insight into where models attend, it lacks structured decision-making and does not link attention patterns to causal reasoning outcomes. Unlike FOCUS, CRM directly tests how removing evidence alters reasoning steps and quantifies step-level reasoning failures. The third theme is Structured Reasoning Augmentation, where methods such as ICoV and RSVP\cite{xu2025rsvp} augment reasoning with structured pipelines. RSVP decomposes images into regions and applies multimodal CoT but depends heavily on segmentation accuracy. ICoV breaks questions into subquestion–answer pairs, ranks visual regions with CLIP\cite{clip}, and fine-tunes MLLMs to align attention with evidence. However, ICoV depends on external tools like GPT-4o and CLIP, making it partially reliant on upstream model quality. Unlike RSVP and ICoV, CRM is training-free and exposes reasoning weaknesses directly through causal intervention on visual inputs.

     The fourth theme is Localization-Enhanced Pretraining covering models like CLOC\cite{chen2024cloc}. CLOC is a Localization-Enhanced Pretraining method that improves the localization capability of CLIP by complementing CLIP with region-text contrastive loss and modules. Unlike CLIP, CLOC scales to billions of annotated images and generates spatially-localized region embeddings. This can be a replacement for CLIP in MLLMs, especially for tasks that requires grounding and fine-grained region reasoning. However, CRM probes the reasoning process by testing the difference masking an image makes while CLOC improves CLIP's localization ability.
    
    Apart from the four themes described above, there are faithfulness evaluation frameworks like Causal Diagnosticity\cite{llmfaithfullness2025}, which introduces a structured way to test whether explanation methods capture a model's reasoning process. This method emphasizes evaluation of explanations and leverages model-editing methods to generate faithful-unfaithful explanation pairs. The benchmark includes fact-checking, analogy, object counting, and multi-hop reasoning. It is evaluated with faithfulness metrics like post-hoc explanations and Chain-of-thought methods. It was found that diagnostic performance is different across different tasks and models. While the focus is on text-based LLM explanations, this method aligns closely with CRM's motivations.




\section{Methods and Results}
\begin{figure*}[t]
    \centering
    \includegraphics[width=0.8\textwidth]{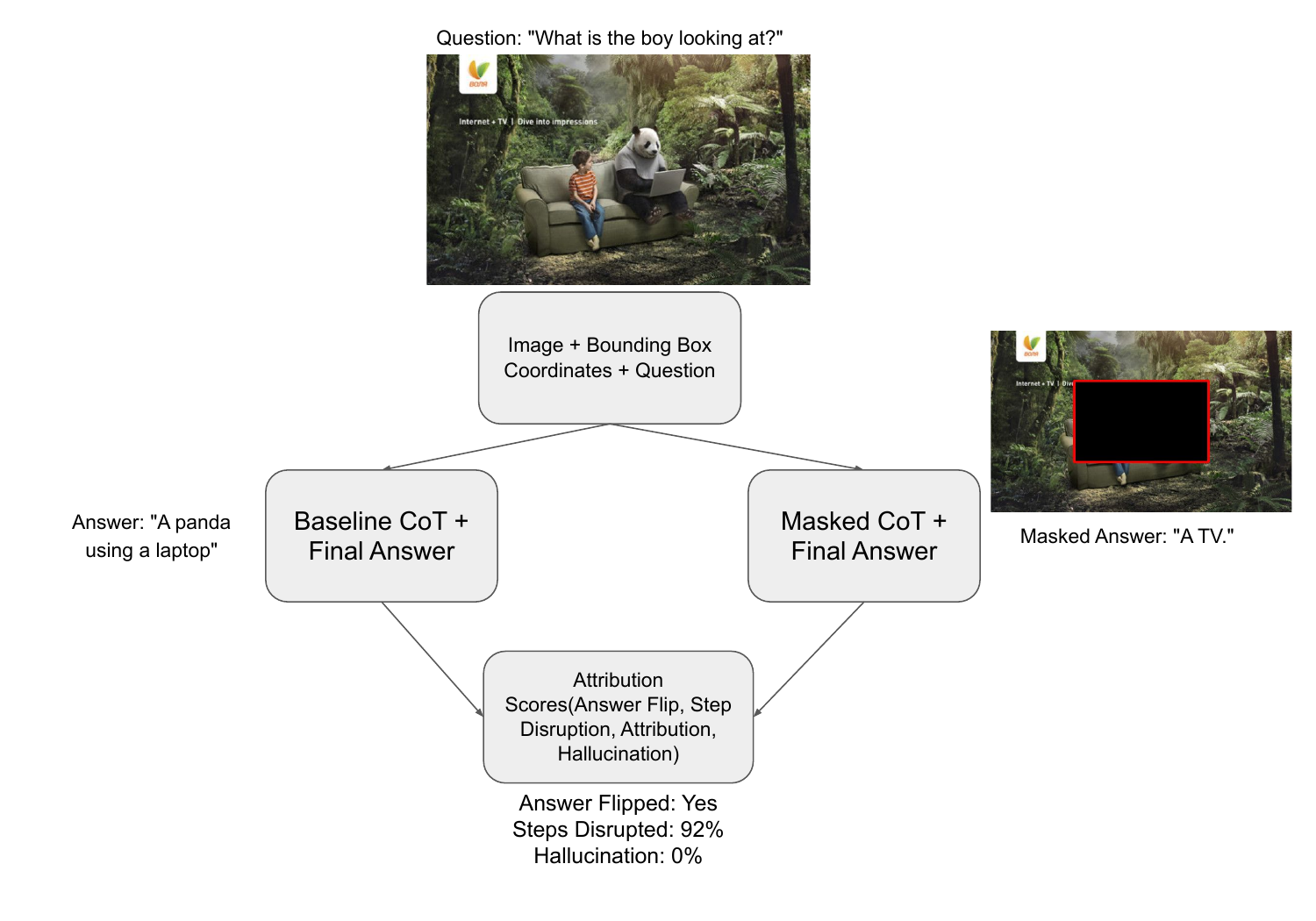}
    \caption{Architecture for Contrastive Region Masking (CRM), which contrasts original and masked chains of thought (CoT) and their final answers to assess the importance of visual evidence in reasoning.}
    \label{architecture}
\end{figure*}

\begin{table}[t]
\centering
\scriptsize
\setlength{\tabcolsep}{6pt}
\renewcommand{\arraystretch}{1.15}
\caption{Attribution steps for Fig.~\ref{architecture}: comparison of Original vs.\ Masked chain-of-thought (CoT).}
\label{tab:attribution_steps}
\begin{tabularx}{\linewidth}{@{}lllYY@{}}
\toprule
\textbf{Step} & \textbf{Status} & \textbf{Cue} & \textbf{Original (summarized)} & \textbf{Masked (summarized)} \\
\midrule
0  & Unchanged   & Scene          & Young boy on sofa in lush forest. & Boy on couch in dense jungle. \\
1  & Modified    & Secondary obj. & Panda on sofa using a laptop. & Large black rectangle in front of couch; TV-like shape. \\
2  & Modified    & Gaze/posture   & Boy looks at panda. & Posture suggests focus on something in front. \\
3  & Modified    & Background   & Background is a dense forest & A logo for a company along with text "Internet + TV | Dive" \\
4  & Modified    & Text/logo      & “Internet + TV | Dive into impressions” with logo. & Lush jungle evokes nature/escape/relaxation. \\
5  & Modified    & Setting        & Forest implies natural/remote environment. & Couch implies relaxation/entertainment context. \\
6  & Modified    & Symbolism      & Panda connotes nature/peace/tranquility. & Large black rectangle strongly suggests a TV screen. \\
7  & Modified    & Device         & Laptop implies tech/info/entertainment access. & Logo/services indicate the image is an advertisement. \\
8  & Modified    & Attention      & Direct looking $\Rightarrow$ attention on object/being. & People typically face the TV when watching. \\
9  & Modified    & Ad framing     & Text frames an internet/TV services ad; “immersive.” & Given posture + rectangle placement, likely looking at a TV. \\
10 & Modified    & Juxtaposition  & Boy’s gaze at panda with laptop $\Rightarrow$ observing activity. & Jungle vs.\ TV suggests nature–tech contrast / access from anywhere. \\
11 & Modified    & Message        & Nature (forest, panda) + tech (laptop) $\Rightarrow$ access even in remote places. & Ad conveys entertainment/connectivity anywhere; \emph{conclusion:} boy looks at a TV. \\
12 & Disappeared & Mask effect    & Ad message: “Dive into impressions”; boy attends to panda–laptop interaction; \emph{conclusion:} boy looks at panda using laptop (tech immersion). & Not applicable—the masking fully disappeared. \\
\bottomrule
\end{tabularx}
\end{table}

\subsection{Data and Annotation Statistics}
Our tests are conducted on 1,611 VisArgs benchmarks. VisArgs consists of image–question pairs with relevant and irrelevant region annotations, aimed at selective evaluation of visual reasoning. We sampled a balanced subset of both the annotations across topics and difficulty levels to ensure the highest diversity in visual content and reasoning complexity. 

\subsection{Pre-Masking CoT Generation}
For each LLM that we use, we first get baseline CoT generations where we run CoT prompting on full unmasked images to log complete reasoning traces and answers as baseline output. Each question is paired with a bounding box that indicates a specific visual region. For these generations, we use a temperature of 0.2 for the CoT reasoning stage and a temperature of 0.0 for the final answer stage. Each question outputs a CoT answer and a final answer for the unmasked region and bounding box. To determine how much the model reasoning and analysis change with masked images, we evaluate and compare the baseline CoT with the CoT traces in masked images with attribution scores. Figure \ref{architecture} illustrates the overall pipeline of the process, while Table \ref{tab:attribution_steps} details the attribution steps of the baseline generation and the corresponding masking generation.

\subsection{CoT Prompts}

\paragraph{Baseline CoT Generation:}
We use the following Chain-of-Thought (CoT) prompt template to elicit explicit visual and commonsense reasoning, followed by a concise final conclusion.

\begin{promptbox}[CoT Prompt Template]
Think step by step to answer the given question about the image, and explain what message or idea it is trying to convey.
Use your observations and commonsense knowledge.

1) Visual Premises (VP): Identify and describe all key visual elements relevant to the question.
2) Commonsense Premises (CP): For each VP, explain the typical meaning, function, or implication of the element based on commonsense knowledge.
3) Inference/Conclusion (IC): Combine VP and CP to reason explicitly about what the image is showing and what it implies in relation to the question.
4) Final Conclusion: Provide a final answer to the question based on the reasoning.

Example
Image: A cup is pouring tea into a brain. Below the brain, there is a loading bar with the words "Loading CreaTEAvity".
Question: What is being poured into the brain?
Reasoning:
- VP1: A cup is pouring a brown liquid into a brain.
- VP2: The brain is partially filled with the liquid.
- VP3: There is a progress bar labeled "Loading CreaTEAvity."
- CP1: Brains symbolize intellect and creativity.
- CP2: Brown liquids like tea or coffee are associated with mental stimulation.
- CP3: A progress bar indicates an ongoing process or enhancement.
- IC1: The brown liquid represents tea being used to metaphorically enhance mental creativity.
- IC2: The progress bar reinforces the idea of creativity being gradually boosted.
Final Conclusion: Tea is being poured into the brain, suggesting it enhances creativity.

Question: {question}
\end{promptbox}

\paragraph{Final Answer generation (No CoT):}
For final answer generation that requires only a succinct response without revealing intermediate reasoning, we use the following answer template.

\begin{promptbox}[Answer Template]
Provide a final short answer to the question, based only on the image and bounding box context.
Do not include step-by-step reasoning.
Question: {question}

\end{promptbox}

\paragraph{Masking CoT Generation:}
We use the same general CoT prompt in the masking function compared to the baseline CoT prompt. For the answer generation that require a response, we use:

\begin{promptbox}[Answer Template]
Provide the final short answer based only on the masked image and context. Do not include step-by-step reasoning.
Question: {question}
\end{promptbox}

\subsection{Region Masking on CoT}
For bounding box evaluation, we apply a mask to regions that are explicitly identified as important in the given dataset. These important regions are defined according to the dataset’s annotations. After masking these specified regions, we reintroduce the original question to ensure consistency between masked and unmasked runs. We use the same temperature values as in the pre-masking CoT generation phase to guarantee consistency in inference behavior. The model then processes the modified image to produce a masked chain-of-thought (CoT) trace and final answer, which are compared against the baseline outputs.

\subsection{Attribution Scores}
In this stage, we assess how masking affects the model's reasoning and final predictions. First, we identify which reasoning steps and final answers are disrupted by the masking. We then measure the degree to which the baseline CoT and final answer semantically differ from the masked CoT and final answer. All reasoning steps that differ are marked. Semantic steps are characterized by hallucinated reasoning steps, disappearance of reasoning steps, or changes in the final answer. We observe the following evaluation metrics:
\begin{itemize}
  \setlength\itemsep{0pt}
  \setlength\parskip{0pt}
  \setlength\parsep{0pt}
  \item \textbf{Answer Flipped}:~Measures overall task sensitivity to visual region removal.
  \item \textbf{Step Disrupted}:~Captures which reasoning steps change significantly, counting the entire image as disrupted if any step is modified
  \item \textbf{Region Attribution}:~Evaluates alignment between GT regions and actual reasoning dependencies.
  \item \textbf{Similarity}: We use Sentence-BERT all-MiniLM-L6-v2 \cite{reimers2019sentencebert} to calculate semantic similarity between the reasoning steps for original images and the masked image. For this, we set a threshold of 0.80 and use it to measure the step-disrupted metric to capture moderate changes while still reflecting consistency. We use a similar method with a threshold of 0.90 to measure the answer-flipped metric to focus on the correctness of the final outcome, where even a small deviation can change the final answer semantics.
  \item \textbf{Hallucination}:~Surfaces when models continue referring to missing content, counting the entire image as hallucinated if any step introduces spurious information.
\end{itemize}

The metric results are in Table~\ref{tab:results_evaluation}.
Table~\ref{tab:results} presents the disruption analysis of visual reasoning steps under masked region perturbations. We evaluate three metrics—Answer Flip Rate, Step Disruption Rate, and Hallucination Rate—for Gemini-1.5-Flash, GPT-4o, Qwen-2.5-VL-7B-Instruct, and Llama-3.2-90B-Vision-Instruct. We selected models of varying sizes and types to understand how these factors influence their responses. The models exhibit contrasting behaviors under masked region perturbations. For the Answer Flip Rate, Gemini-1.5-Flash records a value of 58.78\%, while GPT-4o shows a much higher rate of 74.74\%, and Llama-3.2-90B-Vision-Instruct at 75.72\%, and Qwen-2.5-VL-7b-Instructs shows the highest flip rate at 85.72\%, indicating that Qwen-2.5-VL-7B-Instruct is the most sensitive to regional masking, GPT-4o and Llama-3.2-90B-Vision-Instruct are moderately sensitive, and Gemini-1.5-Flash is the most stable. In terms of the Step Disruption Rate, Gemini-1.5-Flash shows 79.08\% of reasoning steps disrupted compared to 92.86\% for GPT-4o, 95.59\% for Qwen-2.5-VL-7B-Instruct, and 93.73\% for Llama-3.2-11B-Vision-Instruct, suggesting that while all models are heavily affected by perturbations, Gemini-1.5-Flash retains the most step-level coherence, Llama-3.2-90B-Vision-Instruct and GPT-4o are heavily disrupted, and Qwen-2.5-VL-7B-Instruct suffers the most. However, when considering the Hallucination Rate, Qwen-2.5-VL-7B-Instruct hallucinates the least (26.57\%), Gemini-1.5-Flash slightly more (30.60\%), Llama-3.2-90B-Vision-Instruct higher still (49.97\%), and GPT-4o the most (35.51\%). This shows that Qwen-2.5-VL-7B-Instruct gives the most grounded responses, Gemini-1.5-Flash is moderately grounded, GPT-4o hallucinates more often, and Llama-3.2-90B-Vision-Instruct is the least grounded overall. 

GPT-4o sometimes explicitly refused to perform the masked CoT reasoning, preferring to state its limitations rather than generate unsupported content. Qwen-2.5-VL-7b-Instruct, while having stable answers, occassionaly inserted random letters within the CoT reasoning when part of the image were masked. Together, these results highlight a four-way trade-off: Gemini-1.5-Flash is the most stable (lowest flip and disruption rates), though it hallucinates moderately. Qwen-2.5-VL-7B-Instruct is the most grounded with lowest hallucination but also exhibits the highest disruption and flip rates. GPT-4o is highly sensitive to masking, with frequent answer flips but refuses unsupported reasoning rather than hallucinating, which reflects different robustness profile. Llama-3.2-90B-Vision-Instruct suffers with both high disruption and hallucination, making it least reliable. Also importantly, the tight error margins (<1.3\%) shows that these differences are robust across samples. CRM thus surfaces systematic weaknesses that would be invisible if only the final answer was measured. 

\begin{figure}[h]
    \centering
    \subcaptionbox{}{\includegraphics[width=0.35\textwidth]{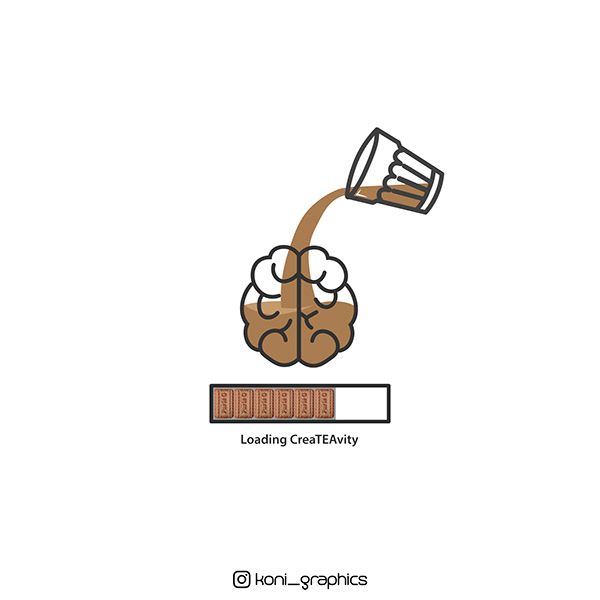}\label{fig:1a}} \qquad
    \subcaptionbox{}{\includegraphics[width=0.35\textwidth]{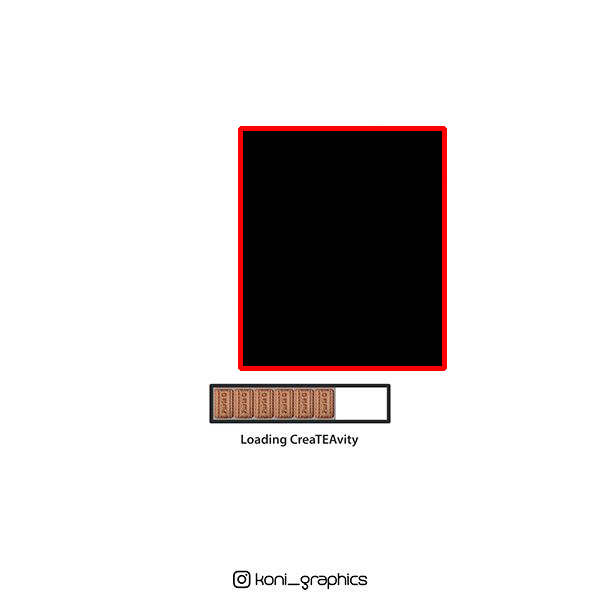}\label{fig:1b}}
    \caption{Brain\_Loading\_Tea: (a) Original Image, (b) Masked Image.  Question: "What is being poured into the brain in the image?" A case where Gemini-1.5-Flash, GPT-4o, Qwen-2.5-VL-7b-Instruct, and Llama-3.2-90B-Vision-Instruct models diverge in Answer Flipped and Region Attribution, showcasing the different behavior of the models as a result of masking (Table~\ref{tab:results}).}
    \label{fig:example}
\end{figure}
\begin{figure}[h]
    \centering
    \subcaptionbox{}{\includegraphics[width=0.20\textwidth]{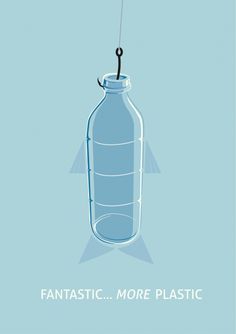}\label{fig:2a}} \qquad
    \subcaptionbox{}{\includegraphics[width=0.20\textwidth]{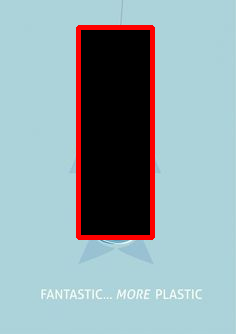}\label{fig:2b}}
    \caption{Fish\_Container: (a) Original Image, (b) Masked Image. Question: "What type of container is shown hanging from the fishing hook?" A case where CoT steps were disrupted across all the models, but Gemini-1.5-Flash and  Llama-3.2-90B-Vision-Instruct showcase hallucination as well after masking (Table~\ref{tab:results}).}
    \label{fig:example3}
\end{figure}





\begin{table}[t]
\centering
\scriptsize
\caption{Comparison of evaluation metrics across models.\textbf{Qwen-2.5-VL-7B-Instruct} shows the highest Answer Flip Rate and Step Disruption, indicating stronger sensitivity to masking. \textbf{Llama-3.2-90B-Vision-Instruct} has the highest hallucination rate.}
\label{tab:results_evaluation}
\begin{tabular}{lcccc}
\toprule
\textbf{Metric} & \textbf{Gemini-1.5-Flash} & \textbf{GPT-4o} & \textbf{Qwen-2.5-VL-7B-Instruct} & \textbf{Llama-3.2-90B-Vision-Instruct} \\
\midrule
Answer flip rate   & $58.78 \pm 1.23\%$ & $74.74 \pm 1.08\%$ & $\mathbf{85.72 \pm 0.87\%}$ & $75.72 \pm 1.07\%$ \\
Step disruption    & $79.08 \pm 1.01\%$ & $92.86 \pm 0.65\%$ & $\mathbf{95.59 \pm 0.51\%}$ & $93.73 \pm 0.60\%$ \\
Hallucination      & $30.60 \pm 1.15\%$ & $35.51 \pm 1.19\%$ & $26.57 \pm 1.10\%$ & $\mathbf{49.97 \pm 1.25\%}$ \\

\bottomrule
\end{tabular}
\end{table}

\begin{table}[!t] 
\centering
\scriptsize
\setlength{\tabcolsep}{5pt}
\renewcommand{\arraystretch}{1.15}
\caption{Disruption analysis of visual reasoning under masked region perturbations. 
For each image–model pair, the table shows the ground-truth reasoning step (GT Step), the masked region, whether the step was disrupted, the answer flipped, or hallucination occurred, and region attribution (bucketed as incorrect, correct, and partial), as well as similarity.}
\label{tab:results}
\begin{tabular}{llp{4.5cm}p{4cm}}
\toprule
\textbf{Image} & \textbf{Model} & \textbf{GT Step / Masked Region} & \textbf{Details} \\
\midrule
Brain\_Loading\_Tea & Gemini-1.5-Flash &
Step~1 (VP1 ``glass pouring tea'') / Box (glass$\to$brain) &
-- Step disrupted: Yes \newline
-- Answer flipped: Yes (Tea $\to$ Creativity) \newline
-- Similarity: $\approx$0.0 \newline
-- Region attribution: Incorrect \newline
-- Hallucination: No \\
\addlinespace[0.3em]
Brain\_Loading\_Tea & GPT-4o &
Step~1 (VP1 ``brown liquid being poured'') / Box (glass$\to$brain) &
-- Step disrupted: No \newline
-- Answer flipped: No \newline
-- Similarity: $\approx$1.0 \newline
-- Region attribution: Correct \newline
-- Hallucination: No \\
\addlinespace[0.3em]
Brain\_Loading\_Tea & Qwen-2.5-VL-7B-Instruct &
Step~1 (VP1 ``brown liquid poured from cup'') / Box (glass$\to$brain) &
-- Step disrupted: Yes\newline
-- Answer flipped: No  \newline
-- Similarity: $\approx$0.0 \newline
-- Region attribution: Partial \newline
-- Hallucination: Yes \\
\addlinespace[0.3em]
Brain\_Loading\_Tea & Llama-3.2-90B-Vision-Instruct &
Step~1 (VP1 ``brown liquid poured into brain'') / Box (glass$\to$brain) &
-- Step disrupted: Yes\newline
-- Answer flipped: Yes (Tea $\to$ Coffee) \newline
-- Similarity: $\approx$0.0 \newline
-- Region attribution: Partial\newline
-- Hallucination: Yes\\
\addlinespace[0.3em]
Soda\_Plug\_CocaCola & Gemini-1.5-Flash &
Step~1(plug+bottles as prongs)/Box (plug) &
-- Step disrupted: Yes \newline
-- Answer flipped: No \newline
-- Similarity: $\approx$0.82 \newline
-- Region attribution: Partial \newline
-- Hallucination: No \\
\addlinespace[0.3em]
Soda\_Plug\_CocaCola & GPT-4o &
Step~1(plug+labels) / Box (plug) &
-- Step disrupted: Yes \newline
-- Answer flipped: No \newline
-- Similarity: $\approx$0.0 \newline
-- Region attribution: Incorrect \newline
-- Hallucination: No \\
\addlinespace[0.3em]
Soda\_Plug\_CocaCola & Qwen-2.5-VL-7B-Instruct &
Step~1 VP1 ``black rectangular podium with Coca-Cola bottles'') / Box (plug) &
-- Step disrupted: Yes\newline
-- Answer flipped: No \newline
-- Similarity: $\approx$0.0 \newline
-- Region attribution: Incorrect \newline
-- Hallucination: Yes \\
\addlinespace[0.3em]
Soda\_Plug\_CocaCola & Llama-3.2-90B-Vision-Instruct &
Step~1 (VP1 ``black object resembling fridge or plug'') / Box (plug) &
-- Step disrupted: Yes\newline
-- Answer flipped: No \newline
-- Similarity: $\approx$0.45 \newline
-- Region attribution: Partially incorrect \newline
-- Hallucination: No \\
\addlinespace[0.3em]
Fish\_Container & Gemini-1.5-Flash &
Step~1 (VP2 ``transparent bottle hanging'') / Box (bottle) &
-- Step disrupted: Yes \newline
-- Answer flipped: Yes (Bottle $\to$ Plastic bag) \newline
-- Similarity: $\approx$0.62 \newline
-- Region attribution: Partial \newline
-- Hallucination: Yes \\
\addlinespace[0.3em]
Fish\_Container & GPT-4o &
Step~1 (VP1 ``clear plastic bottle'') / Box (bottle) &
-- Step disrupted: Yes \newline
-- Answer flipped: No \newline
-- Similarity: $\approx$0.0 \newline
-- Region attribution: Incorrect \newline
-- Hallucination: No \\
\addlinespace[0.3em]
Fish\_Container & Qwen-2.5-VL-7B-Instruct &
Step~1 (VP1 ``rectangular shaped object hanging from hook'') / Box (bottle) &
-- Step disrupted: Yes\newline
-- Answer flipped: Yes\newline
-- Similarity: $\approx$0.0 \newline
-- Region attribution: Partial\newline
-- Hallucination: No \\
\addlinespace[0.3em]
Fish\_Container & Llama-3.2-90B-Vision-Instruct &
Step~1 (VP1 ``black rectangular object hanging from string'') / Box (bottle) &
-- Step disrupted: Yes\newline
-- Answer flipped: Yes\newline
-- Similarity: $\approx$0.0\newline
-- Region attribution: Partial\newline
-- Hallucination: Yes\\
\bottomrule
\end{tabular}

\end{table}


\subsection{Ablation Study}


To systematically assess model robustness, we perform a dual-ablation study evaluating GPT-4o, Gemini-1.5-Flash, Qwen-2.5-VL-7B-Instruct, and Llama-3.2-90B-Vision-Instruct under both specific masking and random masking.

In the random masking, irrelevant regions were selected by randomly picking image regions that were at least five percent away from the original masked region to avoid overlap. In specific masking, we measure the model's robustness to masking important regions provided in the \cite{zhang2024selective} dataset. This lets us see how sensitive models are to missing information and how robust they are to noise. 


 We consider three core metrics for this study: \textbf{Hallucination Rate}, \textbf{Step Disruption Rate}, and \textbf{Answer Flip Rate}. We perform a detailed comparative analysis of specific masking and random masking in GPT-4o, Gemini-1.5-Flash, Qwen-2.5-VL-7B-Instruct, and Llama-3.2-90B-Vision-Instruct using the three metrics. The results of the ablation study are presented in Table~4 and Table~5. 
 


\begin{table}[!t]
\centering
\scriptsize
\caption{Random Masking results for models. Key results include Qwen achieving lowest hallucination and high answer flip rate. Gemini minimizes answer flips and lower step disruption but has hallucination struggles. Bold percentages show some of the key results.}
\label{tab:random_masking}
\begin{tabular}{lccc}
\toprule
\textbf{Model} & \textbf{Hallucination Rate (\%)} & \textbf{Step Disruption Rate (\%)} & \textbf{Answer Flip Rate (\%)} \\
\midrule
GPT-4o              & $33.27 \pm 1.17\%$ & $92.79\pm 0.64\%$ & $58.03 \pm 1.23\%$ \\
Gemini-1.5-Flash    & \textbf{\boldmath{$52.64 \pm 1.24\%$}} & \textbf{\boldmath{$79.02 \pm 1.01\%$}} & \textbf{\boldmath{$26.65 \pm 1.12\%$}} \\

Qwen-2.5-VL-7B-Instruct    & $29.74 \pm 1.14\%$ & $95.59 \pm 0.51\%$ & \textbf{\boldmath{$75.54 \pm 1.07\%$}} \\

Llama-3.2-90B-Vision-Instruct & $41.59 \pm 1.23\%$ & $93.73\pm 0.60\%$ & $55.37\pm 1.24\%$ \\
\bottomrule
\end{tabular}
\end{table}

\begin{table}[t]
\centering
\scriptsize
\caption{Specific Masking results for models. Key results include Qwen behaving similar to random masking having lowest hallucination and higher answer flip rates. Gemini has the lowest step disruption rate. Bold percentages show some of the key results.}
\label{tab:specific_masking}
\begin{tabular}{lccc}
\toprule
\textbf{Model} & \textbf{Hallucination Rate (\%)} & \textbf{Step Disruption Rate (\%)} & \textbf{Answer Flip Rate (\%)} \\
\midrule
GPT-4o& $35.51 \pm 1.18\%$ & $92.86 \pm 0.65\%$ & $74.74 \pm 1.08\%$ \\
Gemini-1.5-Flash& $30.60 \pm 1.15\%$ & \textbf{\boldmath{$79.08 \pm 1.01\%$}}

& $58.78 \pm 1.23\%$ \\
Qwen-2.5-VL-7B-Instruct      & $26.57 \pm 1.10\%$ & $95.59 \pm 0.50\%$ & \textbf{\boldmath{$85.72 \pm 0.87\%$}} \\

Llama-3.2-90B-Vision-Instruct & $49.97 \pm 1.25\%$ & $93.73 \pm 0.55\%$ & $75.72 \pm 1.07\%$ \\
\bottomrule
\end{tabular}
\end{table}

\textbf{Hallucination Rate:}

 GPT-4o maintains a relatively low hallucination rate across both settings, rarely introducing unsupported reasoning or invented content, even when input is masked. Gemini-1.5-Flash, by contrast, exhibits a high hallucination rate (52.64\%) under random masking and remains notably prone to hallucination under specific masking (30.60\%), indicating increased risk of spurious outputs, especially under missing evidence. Qwen-2.5-VL-7B-Instruct remains the most grounded with lowest hallucination rate (26.65\%), suggesting strong evidence-based reasoning even when perturbed. Llama-3.2-90B-Vision-Instruct shows higher hallucination (41.59\% random, 49.97\% specific), reflecting weaker resistance to masking. 
\newline
\newline
\textbf{Step Disruption Rate:}

 All models showed high sensitivity to masking at the reasoning-step level, but extent varied. GPT-4o is markedly affected with 95.59\% disruption under both random and specific masking. Gemini-1.5-Flash is less affected with disruption rates of 79.02\% under random masking and 79.08\% under specific masking. Qwen-2.5-VL-7B-Instruct exhibits the highest disruption rates of 95.59\% for both random and specific masking. Llama-3.2-90B-Vision-Instruct shows similar disruption to GPT-4o for both specific and random masking. 
 \newline
 \newline
 \textbf{Answer Flip Rate:}

 The final answer stability shows clear model divergences. GPT-4o displays high answer flip rates (58.03\% random, 74.74\% masked), indicating that its final predictions are highly variable with perturbed or incomplete input. Gemini-1.5-Flash is more stable under general conditions (26.65\%) but its answer flip rate rises sharply to 58.78\% with visual region masking, reflecting reduced answer stability in the absence of key information. Qwen-2.5-VL-7B-Instruct is the most unstable with both flip rates exceeding 75\% (75.54\% random, 85.72\% specific), showing its severe fragility in its final answers. Llama-3.2.90B-Vision-Instruct also exhibits high instability (55.37\% random, 75.72\% specific), showing weakness when essential parts of image is removed. 
    \newline
    \newline
    \textbf{Model Behaviors:}

 In environments with general linguistic noise or synthetic perturbation, GPT-4o maintains solid factual correctness but is easily destabilized. Gemini-1.5-Flash prioritizes reasoning stability shown from lower disruption rates and lower answer flip rates. Qwen-2.5-VL-7B-Instruct shows the least hallucination rate but is extremely instable in both reasoning steps and answers. Finally, Llama-3.2-90B-Vision-Instruct has moderate hallucination rates and a higher rate for the models present, it also has high step disruption and flip rates as well, a more similar behavior to GPT-4o but weaker factual grounding. 
    \newline 
    \newline
    \textbf{Robustness Profile:}

 The aggregate results across both ablations indicate a trade-off between factual grounding and reasoning/answer stability. GPT-4o delivers consistently fact-based outputs with relatively low hallucination rates yet suffers significant instability in both intermediate reasoning and final answer when inputs are incomplete or noisy. Gemini-1.5-Flash, in contrast, prioritizes stability and consistency, yet this comes with a higher likelihood of generating unsupported or spurious content, particularly as input completeness degrades. Qwen-2.5-VL-7B-Instruct is the most grounded, yet also the highest step disruption rates as well as answer flip rates. Llama-3.2-90B-Vision-Instruct is more unstable overall, reflecting high vulnerability in both reasoning steps and final answers. 

These findings characterize the fundamental strengths and weaknesses of each model class, offering targeted insights for future research on balancing stability, robustness, and grounding in multimodal and language models.

\section{Limitations}
While our study advances step-level visual reasoning, it has four key limitations.  The study relies on the clean, well-annotated VisArgs dataset, which may not reflect noisy real-world images very well. CRM is a diagnostic, not a corrective study, so it identifies flaws but doesn't solve the reasoning problems. The scope of the study is also restricted to static images, with richer temporal dynamics in video left for future work. Lastly, random masking might not be a sufficient way to find unimportant regions, as they can still overlap with semantically relevant content, so further analysis is needed to find unimportant regions.



\section{Conclusion and Future Work}
In this work, we systematically masked individual regions of VisArgs images and compared the resulting reasoning traces to the original outputs. This analysis exposed subtle weaknesses in multimodal reasoning, including logical omissions, semantic drift, and answer disruptions, which are often overlooked by traditional final-answer metrics. By revealing step-level causal dependencies, CRM provides a pathway toward more interpretable, transparent, and trustworthy evaluation of MLLMs. By reframing visual benchmarks as diagnostic probes, our framework highlights that future progress should be measured not only by the correctness of answers but also by the fidelity of reasoning traces to visual evidence, emphasizing the importance of models that reason as reliably as they perform.

\end{document}